\documentclass[12pt, a4paper]{article}
\usepackage[font=footnotesize]{caption}
\usepackage[utf8]{inputenc}
\usepackage{graphicx}
\usepackage{authblk}
\usepackage{booktabs}
\usepackage{verbatim}
\usepackage{url}
\usepackage{caption}
\usepackage{subcaption}
\usepackage{indentfirst}
\usepackage{url}
\usepackage{amsmath}
\title{Two Approaches to Supervised Image Segmentation}
\author{Alexandre Benatti and Luciano da F. Costa}

\date{
S\~ao Carlos Institute of Physics - DFCM \protect\\
University of S\~ao Paulo \protect\\
P.O. Box 369, S\~ao Carlos, S.P. \protect\\
13560-970 Brazil \\ \vspace{0.5cm}
\emph{22th Aug., 2023}
}

\begin{document}
\maketitle

\begin{abstract}
Though performed almost effortlessly by humans, segmenting 2D gray-scale or color images into respective regions of interest (e.g.~background, objects, or portions of objects) constitutes one of the greatest challenges in science and technology as a consequence of several effects including dimensionality reduction(3D to 2D), noise, reflections, shades, and occlusions, among many other possibilities. While a large number of interesting related approaches have been suggested along the last decades, it was mainly thanks to the recent development of deep learning that more effective and general solutions have been obtained, currently constituting the basic comparison reference for this type of operation. Also developed recently, a multiset-based methodology has been described that is capable of encouraging image segmentation performance combining spatial accuracy, stability, and robustness while requiring little computational resources (hardware and/or training and recognition time). The interesting features of the multiset neurons methodology mostly follow from the enhanced selectivity and sensitivity, as well as good robustness to data perturbations and outliers, allowed by the coincidence similarity index on which the multiset approach to supervised image segmentation is founded. After describing the deep learning and multiset neurons approaches, the present work develops comparison experiments between them which are primarily aimed at illustrating their respective main interesting features when applied to the adopted specific type of data and parameter configurations. While the deep learning approach confirmed its potential for performing image segmentation, the alternative multiset methodology allowed for enhanced accuracy while requiring little computational resources.
\end{abstract}

\section{Introduction}\label{sec:introduction}

One of the especially important aspects underlying the effective representation and analysis of sensory signals emanates directly from psychophysics (e.g.~\cite{goldstein2021sensation}), more specifically concerning the emphasis that the cognitive systems of living beings place on signal variations, as contrasted to stimuli that are kept constant along time and/or space. In the case of visual perception and cognition (e.g.~\cite{goldstein2021sensation,barlow1972single}), this principle manifests itself particularly in the special importance of \emph{borders} or \emph{contours} (e.g.~\cite{lettvin1959frog,peterhans1991subjective,zhou2000coding,costa2000shape}), which correspond to markedly abrupt variations of visual properties.

The special importance of borders in visual representations can be readily illustrated by the effectiveness of cartoons which, although composed of only a few contours, are able to convey surprisingly rich information regarding not only the identity of the portrayed subjects but also their respective expressions.

The importance of abrupt variations in visual properties, typically gray-level or colors, can be particularly appreciated by considering that the regions of the image where these properties are mostly constant can be understood as being relatively redundant. Indeed, given a rectangular region of uniform gray-level in front of an equally uniform but contrasting background, all the corresponding geometric information can be summarized in terms of the contour of the rectangle.

Observe that it follows from the above considerations that the continuous regions of an image can be treated as being dual to their respective borders, in the sense that they convey the same information. In the present work, we will focus on methods for identifying the latter type of entities, i.e.~regions of images that are characterized by similar gray-levels or colors.

Borders, or alternatively their uniform region duals, play a central role in both natural and artificial image analysis and vision (e.g.~\cite{fu1981survey, pal1993review, zaitoun2015survey, minaee2021image}). Interestingly, the task of identifying these borders or regions corresponds to one of the most difficult challenges in these areas, having continuously motivated a large number of interesting approaches considering a wide range of concepts and methods including but not being limited to: discrete geometry (e.g.~\cite{marchand1999discrete, tobias2002image, mushrif2008color, serra2012mathematical}), statistical physics (e.g.~\cite{romeny2013geometry, winkler2003image}), mathematical physics (e.g.~\cite{szeliski1991physically, kato2012markov, xu2000image, blake2011markov, weickert1998efficient, weickert1998anisotropic}), complex networks (e.g.~\cite{costa2004complex, backes2010shape, camilus2012review}) as well as neuronal networks (e.g.~\cite{egmont2002image, jiang2004som, senthilkumaran2009image}), and more recently, deep learning~\cite{krizhevsky2017imagenet, ronneberger2015u}.

The impressive number of interesting approaches that have been developed for image segmentation confirms not only the importance of this task, but also its challenging nature. Much of the encountered difficulty is related to the multiple sources of noise and imperfections involved in image formation. For example, the gray-level and/or color uniformity of an object or region of an image can become heterogeneous as a result of reflections, shadows, or transparency. In addition, the geometry of the images to be analyzed is typically 2D, as opposed to the 3D geometry of the physical world. These projections from 3D to 2D typically also frequently involve occlusions and overlaps between otherwise uniform regions.

While many interesting approaches to image segmentation have been developed over decades, it was not until the advent of deep learning research (e.g.~\cite{lecun2015deep,krizhevsky2017imagenet,pouyanfar2018survey,arruda2022learning}) that particularly powerful and general segmentation methods were developed that could be applied to effectively segment a wide range of image types. Interestingly, the fields of deep learning and image classification (including segmentation) have become increasingly intertwined and mutually beneficial, leading to significant joint advances. As a result of the above considerations, deep learning-based image segmentation has arguably become the main reference when comparing image segmentation approaches.

Deep learning-based image segmentation is characterized not only by high accuracy and generality, but can also lead to effective knowledge transferring (e.g.~\cite{torrey2010transfer, weiss2016survey, tetteh2020deepvesselnet, da2022analysis}). However, this type of image typically also requires a considerable number of samples accompanied by respective gold standards, as well as substantial software (and/or hardware) resources for respective implementation and operation. Consequently, long periods of time can be typically demanded for training a deep learning systems for image segmentation while relying on relatively massive computational resources. In addition, once one such system is successfully trained, understanding how it works still represents a considerable challenge (e.g.~\cite{nguyen2022interpretable,samek2019towards, zhang2021survey, alvarez2018towards, dong2017improving, meyes2022transparency}). Deep learning systems have also been found to eventually become unstable during the training stage when presented with certain types of untrained patterns (e.g.~\cite{antun2020instabilities, bhadra2021hallucinations, gottschling2020troublesome, xu2023understanding}).

Deep learning developments related to image segmentation aim at developing efficient and accurate models that can segment images effectively. Some of the challenges in this field are dealing with noisy or incomplete data, handling different types of images and objects, and achieving generalization across different domains and scenarios~\cite{goceri2019challenges}. Convolutional Neural Networks (CNNs), consisting of a series of consecutive convolution and pooling layers, are dominant in image segmentation ~\cite{zeiler2014visualizing, nasr2017dense, ronneberger2015u, da2022analysis}. Deep learning has often been used in medical imaging research (e.g.~MRI, CT, X-ray, and ultrasound)~\cite{li2020v, ronneberger2015u, singh20203d,lai2015deep}. 

Recently introduced in the context of multiset operations (e.g.~\cite{da2022multisets, blizard1989multiset, blizard1991development, blizard1989real}), the concept of \emph{coincidence similarity index} (e.g.~\cite{da2021further,costa2022on,costa2023multiset}) is characterized by the implementation of strict comparisons between two discrete or continuous mathematical structures, in the sense of being particularly selective and sensitive. These characteristics have been found to be complemented by many additional desirable properties, including robustness to data perturbations, intrinsic value normalization, and controllability of the sharpness of the implemented comparisons (e.g.~\cite{costa2022on,costa2023multiset}).

Among the several promising applications of the coincidence similarity concept that have been reported so far (e.g.~\cite{costa2022similarity, domingues2022city, benatti2023recovering, benatti2022neuromorphic, da2022autorrelation}), we highlight the possibility of using multiset neurons for implementing the coincidence similarity as a means for achieving effective supervised segmentation of gray-level and color images~\cite{costa2023multiset, da2022supervised}. In particular, this approach incorporates several important features, most of which inherited from the above-observed characteristics, including: (a) just a handful of prototypes (or sample) points are required during the training stage, each being implemented by a respective multiset neuron; b) minimal computational resources (number of neurons) and training time are implied; (c) accurate image segmentation in terms of uniform regions can be typically achieved; (d) possible effective implementation of the involved multiset neurons in analog integrated circuits~\cite{costa2021multiset2}; and (e) good performance has been observed for several types of images.

The present work has the following two main objectives: (i) to present in a relatively comprehensive manner the concept of coincidence similarity-based multiset neuronal networks (henceforward referred to as CS-MNN) and its application to effective supervised image segmentation; and (ii) to compare this methodology with a deep learning-based image segmentation approach based on~\cite{ronneberger2015u, parkhi12a, tensorflow2020docs}.

More specifically, the reported material and developments concerning item (i) above include a complete description of the CS-MNN methodology, the involved parameters, guidelines for their respective setting, as well as several examples considering synthetic and real-world gray-level and color images. 

In addition, experiments are described that compare the CS-MNN and deep learning approaches. In the former case, a relatively small dataset of real-world complex images of granite stone is considered. Given the near impossibility of manually obtaining gold standards for these complex images, and considering the high accuracy provided by the CS-MNN approach, its results are used as a gold standard for this first comparison experiment. As a consequence, this experiment aims to verify the extent to which the deep learning approach can replicate the results obtained by the CS-MNN.

Additional comparison experiments involved a much larger set of synthetic images obtained by using a respectively described generating method which allowed a virtually unlimited number of relatively complex images to be obtained together with their respective gold standard and prototype points. The generation of these images starts by drawing a random field of points, which is then smoothed and thresholded it in order to obtain a binary mask corresponding to the respective gold standard. Two textures (smoothed random fields of points) are then incorporated within the two types of regions of these masks, differing in their average values and spatial scales.

Among the several described results, it has been verified that the deep learning approach confirmed its potential by segmenting the granite images with good accuracy (around 85\% accuracy). After the training stage, the deep learning approach is capable of segmenting the images in a fully independent manner. At the same time, the CS-MNN approach, in addition to yielding highly accurate results, requires a few prototypes or samples (3 in the case of the reported experiment) to be supplied, but the required computational resources are very basic, being limited to the implementation of the respective coincidence similarity operations implied by each of the prototype samples. The deep learning and the CS-MNN performed all segmentation while relying on the same parametric configuration, therefore indicating good generalization among the set of considered granite images.

Similar results were obtained for the additional comparison experiments using synthetic images, with the deep learning approach allowing relatively accurate results, while the CS-MNN method tended to provide more accurate results while requiring a substantially smaller amount of  computational resources.

It should be observed that the comparison reported in the present work is by no means intended to be complete or definitive. Indeed, it should be mainly understood as a study of the main interesting features of both the CS-MNN and deep learning-based approaches to supervised image segmentation approaches. Ultimately, the choice between approaches largely depends on a number of related specific aspects, including resource availability, type of images, desired conceptual framework, and compatibility with other demands, among many other aspects.

This work starts by introducing the fundamental concepts and methodologies employed, such as image analysis principles, balanced accuracy measurements, and the Jaccard and coincidence similarity indices. Subsequently described are concepts of multiset neurons, CS-MNN methodologies adopted for supervised image segmentation, as well as deep learning, and the architecture adopted for image segmentation. The work continues by describing and discussing the results obtained.

\section{Basic Concepts and Methods}\label{sec:Methods}

This section covers some basic concepts of image processing and then describes two approaches to image segmentation by using: CS-MNN, a recently described methodology~\cite{costa2023multiset, da2022supervised}, as well as deep Learning, which constitutes a reference technique in the field.

\subsection{Image Concepts}\label{sec:image}

Among the myriad of data generated every day, visual data tends to be predominant as it underlies and/or is involved in several tasks ranging from biology to astronomy. Basically, an image is a scalar (monochromatic cases including gray-scale images) or vector field (polychromatic cases such as color images) representing some real (or abstract) structure of interest. More traditional types of images include photographs and drawings, as well as those generated by satellites, instrumentation, monitoring, etc.

Images can be defined over 2D or higher dimensional domains (e.g.~\cite{gonzales1987digital,costa2000shape}). The former type is often associated with 2D projections of otherwise 3D geometric spaces, while the latter constitutes a more direct respective representation (e.g.~3D scanning tomography). The basic elements in each of these two cases correspond to the \emph{pixel} (picture element) and \emph{voxel} (volume element). The present work focuses on 2D gray-level and color images.

Given an image, after some pre-processing (e.g.~aimed at reducing noise or other types of enhancements), almost invariably the image needs to be \emph{segmented}
(e,g,~\cite{cheng2001color, haralick1985image, zhang1996survey, pal1993review, zhang1996survey}), which constitutes in identifying the regions of particular interest. Therefore, the task of segmenting an image can be understood as corresponding to a particular instance of pattern recognition.

Typically, the parts of an image to be segmented correspond to separated objects against a background area, or to portions of objects. As already observed in the Introduction section, image segmentation constitutes one of the greatest challenges in the areas of image analysis and computer vision. The segmentation of a given image is typically performed while considering a set or respective relevant features of pixels, which may include gray-level, color, shape, texture, etc. In the present work, we consider the two former types of attributes, but other features can also be adopted.

The representation of the color in the image can take place respectively to several \emph{color systems}, including the RGB and HSV (e.g.~\cite{gonzales1987digital,costa2000shape}), which are the two systems considered henceforth in the present work. The former of these systems represent the color in terms of three components respective to the colors red, green, and blue. In a distinct manner, the HSV systems adopt hue, saturation, and intensity as respective components.

A particularly interesting type of 2D image corresponds to the \emph{binary images}, whose pixels can take only 0 or 1 as values. The assignment of these two values to objects and background is arbitrary, but in this work we understand the object pixels as being represented by the value 1, while the background is represented by the value 0.

Two pixels are henceforth considered 4-adjacent (or 4-neighbors) if they follow one another along the horizontal or vertical directions of a binary image. The 4-neighborhood scheme is adopted henceforth in the present work.

In the case of binary images, the detection of the borders of connected regions can be achieved by taking as a border element every pixel which belongs to an object and that has at least one of its 4 (or 8, in the case of 8-neighborhood) pixels belonging to the background (e.g.~\cite{costa2000shape}).

Binary images can be dilated or eroded respectively to a structuring element (e.g.~\cite{serra2012mathematical}). In the present work, we restrict our attention to the basic $3 \times 3$ cross. In this case, the dilation of the objects in a binary image can be implemented by superimposing one such cross with value 1 onto every pixel belonging to the object of interest in the image. The erosion of an object can be understood as the dilation of the respective remainder of the image.

\subsection{Balanced Accuracy}

In this work, we resource to the Balanced Accuracy ($BA$, e.g.~\cite{garcia2009index}) metric as a means to evaluate the performance of the obtained binary image segmentation. The Balanced Accuracy is defined as the average of the sensitivity and specificity, which are the proportions of true positives and true negatives among the positive and negative instances, which can be mathematically expressed as:
\begin{equation}
   BA =\frac{1}{2}\frac{TP}{(TP+FN)} + \frac{1}{2}\frac{TN}{(TN+FP)},
\end{equation}

where TP = \emph{True Positive}; FP = \emph{False Positive}; TN = \emph{True Negative}; FN = \emph{False Negative}.

This metric is particularly useful when the classes are unbalanced, meaning that one class is more frequent than the other, for instance, when the object is relatively small compared to the background. In this case, a classifier that always predicts the majority class can have a high accuracy, but a low sensitivity and specificity. The Balanced Accuracy can capture both aspects of the performance of the classifier while also taking into account possibly overlooked smaller objects.

\subsection{Jaccard and Coincidence Similarities}

The coincidence measure, which has been considered in recent studies~\cite{costa2022on,da2021further,CostaCCompl}, corresponds to the multiplication between the Jaccard and Interiority indices:
\begin{equation}
   C(\vec{x},\vec{y}) = J(\vec{x},\vec{y}) \ I(\vec{x},\vec{y}).
\end{equation}

The Jaccard index~\cite{Jaccard1, Loet, Jac:wiki, da2021further}, applied to \emph{positive multisets}, can be expressed as:
\begin{equation}
    J(\vec{x},\vec{y}) = \frac{\sum_i min\lbrace x_i, y_i \rbrace}{\sum_i max\lbrace x_i , y_i\rbrace}.
\end{equation}

The Jaccard index applied to multisets is a useful measure of similarity between data sets. It considers both the presence and the frequency of the elements. The interiority~\cite{da2021further}, also known as the overlap~\cite{vijaymeena} is a metric used to evaluate the commutativity between two sets of items. For two multisets with positive entries, the interiority index can be expressed:
\begin{equation}
    I(\vec{x},\vec{y}) = \frac{\sum_i min\lbrace x_i , y_i \rbrace}{ min\lbrace \sum_i x_i ,\sum_i y_i \rbrace}.
\end{equation}

The Coincidence measure can evaluate the similarity between different mathematical entities and implement a rigorous comparison between different structures, such as sets, multisets, vectors, functions, and matrices, among other possibilities~\cite{da2022abrief}.

Optionally, it is possible to include a parameter $D$ in this equation to control the selectivity and sensitivity of the similarity comparisons:
\begin{equation}
    C(\vec{x},\vec{y}, D) = [J(\vec{x},\vec{y})]^D \ I(\vec{x},\vec{y}),
\end{equation}
where $D$ is a non-negative real number. High values of this parameter increase the selectivity of the comparison.

The coincidence approach constitutes a non-dimensional measurement of similarity that has some desirable properties that make it suitable for various applications. One of these properties consists in the fact that the coincidence similarity index values are bound between 0 and 1 (i.e.~$C(\vec{x},\vec{y}) \in [0,1]$), where 0 indicates no similarity and 1 indicates complete similarity. Another property is that it is commutative ($C(\vec{x},\vec{y}) = C(\vec{y},\vec{x})$), which means that the similarity value does not depend on the order of the multisets. A third property is that a multiset is always identical to itself ($C(\vec{x},\vec{x}) = 1$).

In addition, the coincidence similarity has been found~\cite{da2021further,costa2022on,costa2023multiset,CostaCCompl} to implement a comparison between two multisets (or vectors) which is more selective and sensitive than other similarity approaches including, Jaccard and interiority indices, as well as cosine similarity and Pearson correlation coefficient. Other interesting features of the coincidence similarity index include its robustness to limited levels of data perturbations, noise, overlap, and outliers. The coincidence measure can also be generalized to other data structures, including matrices, functions, scalar and vector fields, graphs, etc.

\section{Multiset Neuron-Based Image Segmentation (CS-MNN)}\label{sec:CS-MNN}

The Multiset Neurons-Based Image Segmentation is a methodology proposed recently~\cite{costa2023multiset, da2022supervised} which has been applied to implement supervised image segmentation, while being founded on the coincidence similarity index described in the previous section. This approach involves the selection of a small set of \emph{prototype} or \emph{sample} points ($\mathcal{D}$), which are supposed to provide a suitable description of the region to be identified. 

\begin{figure}
  \centering
     \includegraphics[width=0.4\textwidth]{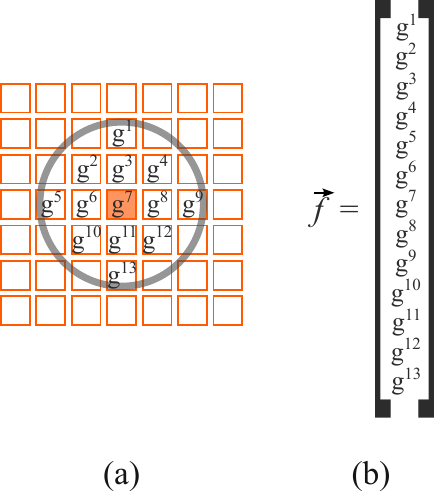}
   \caption{Illustration of the procedure adopted for mapping the gray-level values $g^i$ within the circular region with radius $r=2$ centered at each of the prototype points (a) into a respective $13 \times 1$ feature vector $\vec{f}$ (b). In the case of HSV (or RGB) color images, a $39 \times 1$ single vector would be obtained.}\label{fig:features}
\end{figure}

The gray-level (or color components) values of the pixels contained in a circular window or radius $r$ centered at the coordinate of each of the prototype points are identified and mapped into a respective feature vector (please refer to Fig.~\ref{fig:features}) which is then considered as the weights of a respectively assigned multiset neuron, corresponding to the implementation of a coincidence similarity comparison. These steps constitute the \emph{training stage} of the considered multiset approach to image segmentation.

The \emph{recognition stage} consists of scanning the feature vectors (gray-level or color values organized in the same manner as the weights) of each of the image pixels into the set of multiset neurons. The coincidence similarity values resulting from each of them are then thresholded by $T$ and combined by the logical operation \emph{or}. Thus, a pixel $P$ is considered to belong to the region from which the prototype points have been obtained provided any of the respectively associated multiset neurons yields a high value indicating substantial similarity between the properties (feature vectors) of the point $P$ and the corresponding weight (features of the respective prototype sample).

The effect of the distance $d$ between the considered pixel ($P$) and the closest prototype point ($p \in \mathcal{P}$) can also be taken into account in order to reduce the chances of other portions of the image to be incorporated into the region to be segmented. The relevance of the distance has been defined as:
\begin{equation}
    \mathcal{D}(P,p) = e^{-d a},
\end{equation}
where $a$ is a free parameter controlling the influence of the distance into the segmentation.

The influence of the features and distance can then be combined into the similarity calculation in terms of the respective arithmetic mean:
\begin{equation}
\mathcal{S}(\vec{f}_P,\vec{f}_p) = \frac{C(\vec{f}_P,\vec{f}_p) + \mathcal{D}(P,p)}{2}.
\end{equation}

Table~\ref{tab:parameters} summarizes the parameters involved in the CS-MNN methodology and their respective description.

\begin{table}
\centering
\caption{Parameters of the CS-MNN method and their description.}\label{tab:parameters}
\begin{tabular}{|c|c|}
\hline
\textbf{Parameters} & \textbf{Descrition} \\ \hline
$D$                   & exponent controlling how strict the comparison is  \\
$T$                   & similarity value threshold              \\
$r$                   & window size                             \\
$a$                   & controls the influence of the distance    \\ \hline
\end{tabular}
\end{table}

\begin{figure}
  \centering
     \subfloat[]{\includegraphics[width=0.35\textwidth]{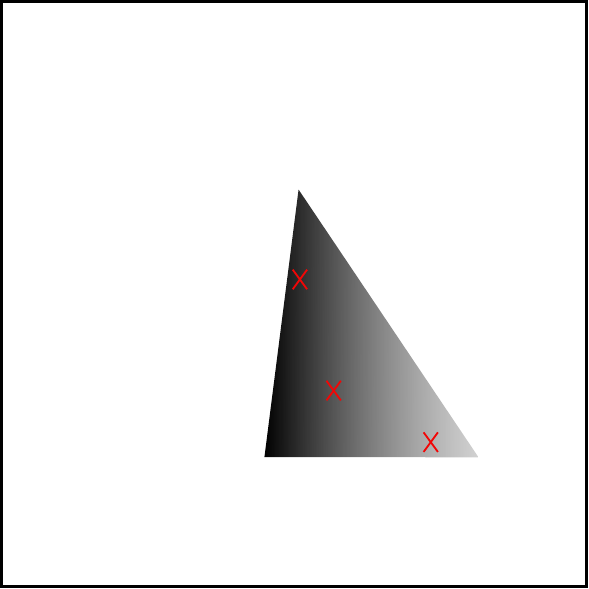}}
      \hspace{0.7cm}
      \subfloat[]{\includegraphics[width=0.35\textwidth]{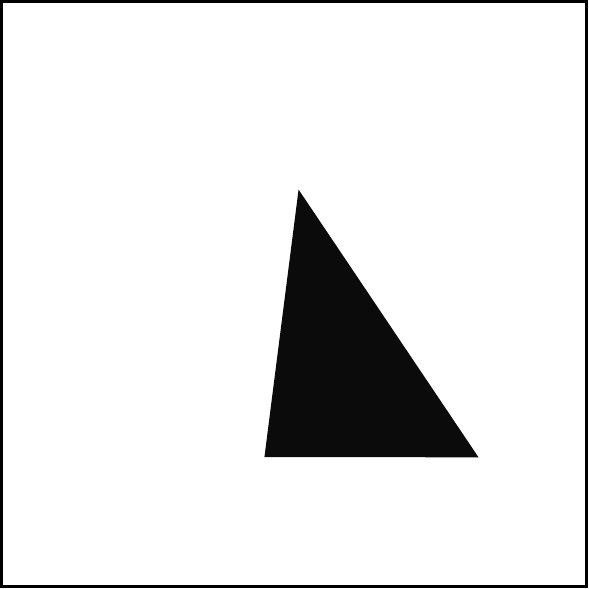}}
     
   \caption{Illustration of the CS-MNN approach to supervised image segmentation adopted in the present work. The object to be segmented corresponds to the triangle, which is respectively sampled in terms of three prototype points (red $\times$'s) providing a suitable representation of the distribution of gray-level values around each of them (points contained in a circular window with radius $r$ centered at the respectively associated prototype coordinates.}\label{fig:fig}
\end{figure}

In order to illustrate the above methodology, which is in this work abbreviated as CS-MNN (for coincidence similarity multiset neuronal network), we now consider the segmentation of the simple gray-level image shown in Figure~\ref{fig:fig}(a), which constitutes of a triangle placed against a background. The objective here is to identify the image pixels that belong to the triangle. 

First, prototype points are chosen (e.g.~interactively) within the triangle, providing a suitable sampling of its properties which, in this case, correspond to gray-level values. The gray-level values of the pixels containing within circular windows with radius $r$ centered at each of the three points are then identified and mapped into a respective vector, therefore corresponding to the weights of three respective multiset neurons respectively associated with each of the three prototypes (samples). The whole image is then scanned into these three neurons, resulting in the segmented region shown in Figure~\ref{fig:fig}(b).

The following guidelines can be taken into account while applying the CS-MNN approach to supervised image segmentation:

\begin{itemize}
   \item If possible, the prototype points should be chosen within the most homogeneous regions, which can be more effectively represented in terms of
   fewer samples;
   \item Particularly heterogeneous regions to be segmented should be samples by a relatively larger number of respective prototype points, chosen in order to sample not only the most typical pattern found in that region, but also other particularly representative variations;
   \item When placing the prototype points, allow for a wide enough inner margin (larger than the parameter $r$) from the respective border, in order to avoid encompassing pixels from other objects or background;
   \item If possible, it can be interesting to select the prototype points in an interactive manner, by clicking the points over a visualization of the respective image to be segmented;
   \item Avoid leaving large gaps in the object or background, in order not to have distance penalization (controlled by the parameter $a$) extending into areas of interest;
   \item When selecting the involved parameters, keep in mind that they may influence one another. For instance, a higher value of $D$ will require a lower threshold $T$. A more comprehensive discussion of the effect of the involved parameters can be found at~\cite{costa2023multiset, da2022supervised}.
\end{itemize}

\section{Deep Learning-Based Image Segmentation}\label{sec:deepsegmentation}

Deep learning-based image segmentation is a technique that uses artificial neural networks to subdivide an image into meaningful regions or objects~\cite{jiang2022intelligent}. Deep learning based-image segmentation has many applications in computer vision~\cite{voulodimos2018deep}, such as medical image analysis~\cite{lai2015deep}, autonomous driving~\cite{fujiyoshi2019deep}, face recognition~\cite{guo2019survey}, and scene understanding, among many other areas. 

Deep learning-based image segmentation methods can achieve high accuracy and robustness by learning from large amounts of data and extracting complex features. Among the most widely used deep learning architectures for image segmentation are Convolutional Neural Networks (CNNs)~\cite{zeiler2014visualizing, ronneberger2015u}, Fully Convolutional Networks (FCNs)~\cite{long2015fully, nasr2017dense}, U-Net~\cite{ronneberger2015u}, V-Net~\cite{li2020v}, and Mask R-CNN~\cite{he2017mask}.

\begin{figure}
  \centering
     \includegraphics[width=1\textwidth]{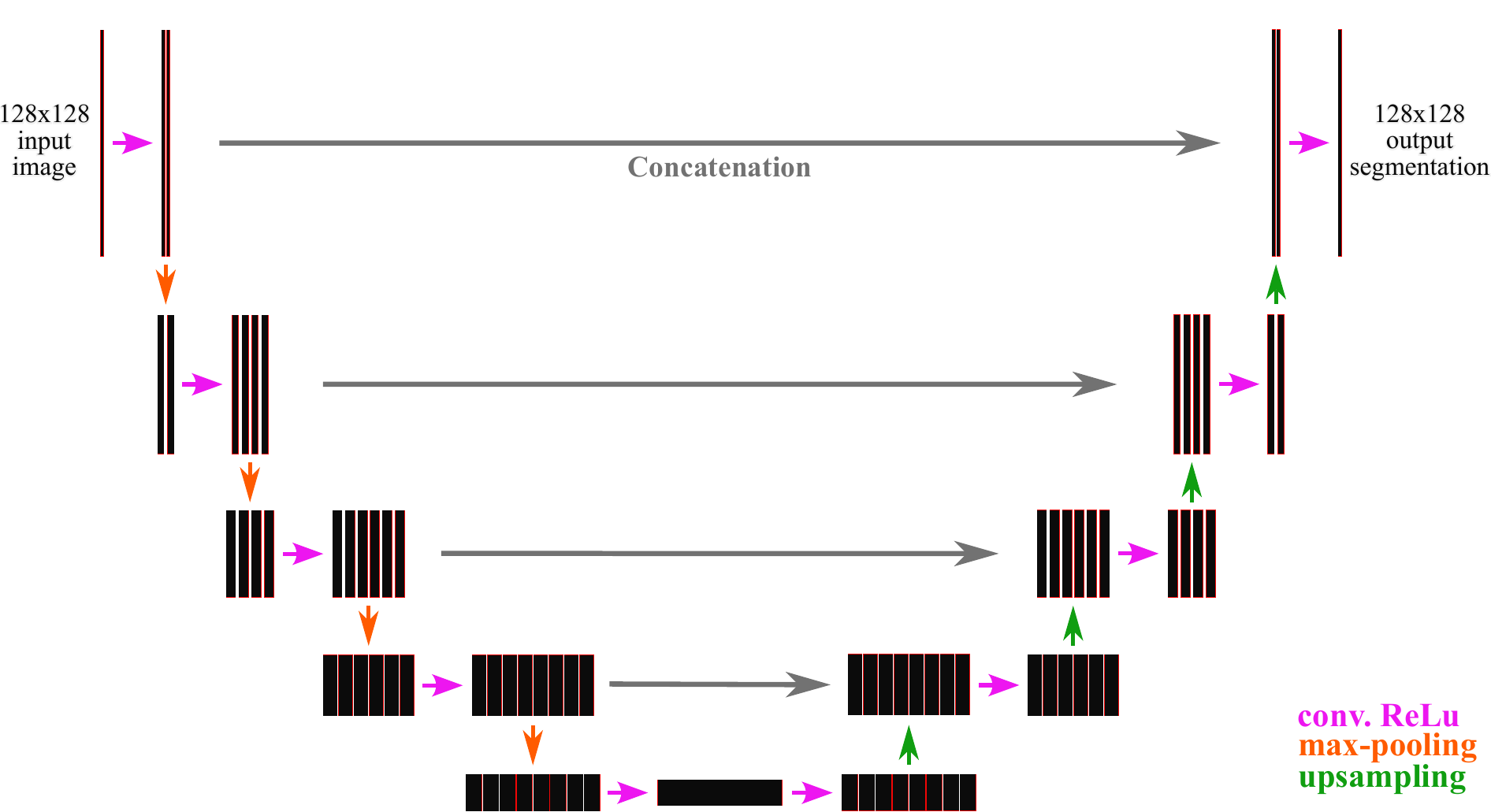}
   \caption{U-net architecture adopted in the current work for image segmentation. The black rectangles indicate the feature maps obtained along different layers of the network. The arrows denote the operations performed on these feature maps, such as convolution, max-pooling, upsampling, and concatenation.}\label{fig:u-net}
\end{figure}

A variation of the U-Net model to perform deep learning-based image segmentation (see Figure~\ref{fig:u-net}) has been adopted. The U-Net model has two parts: a contraction and an expansion. The contraction uses a convolutional neural network model as a feature extractor that outputs a list of feature maps at different resolutions. Each layer in the contraction has repeated convolutions and a max-pooling operation that halves the sample size. The expansion consists of a series of transposed convolution layers that increase the resolution of the feature maps and concatenate them with the corresponding feature maps from the contraction. We considered 128x128 images as the input layer with two channels, corresponding to the class labels of each pixel. This model has been trained by using the Adam optimizer and the sparse categorical cross-entropy loss function~\cite{kingma2014adam}.

\section{Results and Discussion}

In this work, the coincidence approach to supervised image segmentation described in Section~\ref{sec:Methods} is evaluated respectively to four main approaches to be described in the current section: (a) preliminary application to some real-world images of different types; and (b) comparison with a specific deep learning approach respectively to a database containing 45 granite images; (c) comparison considering 200 synthetic images; and (d) comparison respectively to 4500 synthetic images.

The preliminary evaluation (a) aims at illustrating the potential of the reported approach to some different types of real-world images, including beans over a uniform background, a flower vase in a more cluttered environment, and a granite image. Color images (HSV) are considered in all cases, except for the granite image, which has been segmented by taking both its HSV and gray-scale versions into account.

A more systematic evaluation is also reported subsequently in this section, consisting of comparing the coincidence approach with a specific deep learning implementation based on the U-net architecture. A total of 45 colors (HSV) images of distinct granite types, as well as synthetic images, have been considered for segmentation by using the two mentioned approaches.

The present work focuses on the segmentation of raw images, in the sense of involving no pre- or post-processing (e.g.~smoothing, contrast enhancement, adaptive baseline correction, etc.). This choice has been made in order to imply a more demanding segmentation, and also because no pre-processing is often employed in deep learning approaches. However, it should be observed that the CS-MNN (and to some extent also the deep learning approach) can eventually benefit from preliminary smoothing of the image (e.g.~by using Gaussian kernels), as well as adaptive correction of the intensity baseline (e.g.~by using median filtering)~\cite{costa2000shape}.

Enhancements can also be obtained at a post-processing stage. These could include the removal of connected regions with areas smaller than a given threshold, as well as separation of objects, e.g.~by using the opening operation, or filling of voids, e.g.~by using the closing operation (e.g.~\cite{najman2013mathematical,serra2012mathematical,costa2000shape}).

In both the experiments described in this section involving synthetic images, the balanced accuracy of the results is evaluated respectively to gold standard references (binary images).

\subsection{Preliminary Evaluation}\label{sec:Preliminary}

\begin{figure}
  \centering
     \subfloat[]{\includegraphics[width=0.4\textwidth]{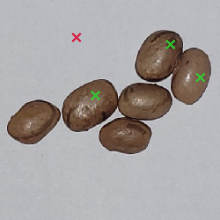}}
     \hspace{0.7cm}
     \subfloat[]{\fbox{\includegraphics[width=0.385\textwidth]{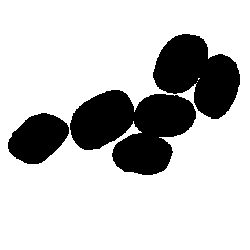}}}
     
     \vspace{0.9cm}
     
     \subfloat[]{\fbox{\includegraphics[width=0.385\textwidth]{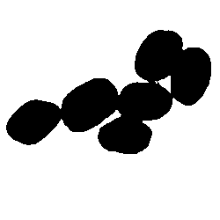}}}
     \hspace{0.7cm}
     \subfloat[]{\includegraphics[width=0.4\textwidth]{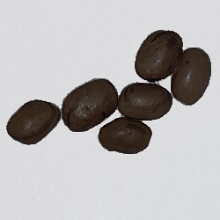}}
     
     \vspace{0.9cm}
     
     \subfloat[]{\fbox{\includegraphics[width=0.385\textwidth]{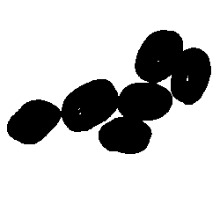}}}
     \hspace{0.7cm}
     \subfloat[]{\includegraphics[width=0.4\textwidth]{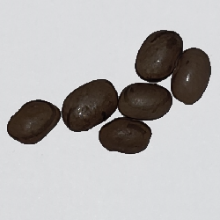}}
     
   \caption{Illustration of the CS-MNN methodology applied to the HSV beans images in (a). Also shown are: the respective gold standard (b), the segmentation result obtained by considering three prototype points (green $\times$'s) sampled from the beans (c), and the superimposition of the latter onto the original image (d). The parameter configuration adopted in this case was $D = 2$, $r = 3$, $T = 0.55$, and $a = 1/ 5000.0$, which yielded a respective accuracy of $BA = 99.06$ \%. The segmentation result obtained for the single background point (red $\times$) is shown in (e), and its superimposition to the original image is presented in (f). The parameter configuration adopted in this case was $D = 1$, $r = 1$, $T = 0.25$, and $a = 1/10.0$, which yielded a respective balanced accuracy of $BA = 99.65$ \%.}

  \label{fig:beans}
\end{figure}

Figure~\ref{fig:beans}(a) presents an HSV image of a set of six beans placed against a uniform background. The objective of the segmentation here is to separate (classify) the beans from the background.

Two alternative approaches are considered in this case, which are specific to distinct manners of choosing the prototype points, namely as: (i) sampled from the objects (beans); and (ii) sampled from the background. The prototype points chosen in each of these two cases are shown respectively in green (three prototype points) and red (one background point).

A larger number of object points has been taken into account in order to provide a more representative sampling of the more heterogeneous image properties characterizing the beans.

Figure~\ref{fig:beans}(b) depicts the gold standard reference image to be considered for quantifying the accuracy of the obtained segmentation. This gold standard has been obtained interactively by human inspection.

By using the gold standard, it has been possible to identify the optimal configuration of the parameters involved in the coincidence-based approach as corresponding to the setting yielding the highest identified balanced accuracy. The optimal parameters identified in cases (i) and (ii) are presented in Table~\ref{tab:beans_parameters}.

\begin{table}
\centering
\caption{Optimal parameters for segmentation of the beans image while taking the prototype samples from the objects and from the background.}\label{tab:beans_parameters}
\begin{tabular}{c|cc|}
\cline{2-3}
                                         & \multicolumn{2}{c|}{\textbf{Optimal Values}}                                                 \\ \hline
\multicolumn{1}{|c|}{\textbf{Parameter}} & \multicolumn{1}{c|}{\textbf{object samples}} & \multicolumn{1}{l|}{\textbf{background samples}} \\ \hline
\multicolumn{1}{|c|}{$T$}                & \multicolumn{1}{c|}{0.55}                   & 0.25                                            \\ \hline
\multicolumn{1}{|c|}{$a$}                & \multicolumn{1}{c|}{1/5,000}                & 1/10                                           \\ \hline
\multicolumn{1}{|c|}{$D$}                & \multicolumn{1}{c|}{2}                      & 1                                              \\ \hline
\multicolumn{1}{|c|}{$r$}                & \multicolumn{1}{c|}{3}                      & 1                                              \\ \hline
\end{tabular}
\end{table}

The optimal segmentation obtained by the coincidence methodology taking into account the case (i) is presented in Figure~\ref{fig:beans}(c), while its superimposition onto the original image is also illustrated in (d). A remarkably accurate result has been obtained, except for the merging of the beans implied by the adopted size of the window $r=3$. If needed, these merging pixels can be readily removed by binary erosion or opening of the obtained results (e.g.~\cite{najman2013mathematical,serra2012mathematical,costa2000shape}).

In the case of using a single background point, the respectively obtained optimal segmentation is shown in Figure~\ref{fig:beans}(e), while (f) presents its superimposition into the original image. The quality of the obtained segmentation resulted again particularly high, though a few incorrect classifications can be observed inside some of the beans as a consequence of the reflexes in Figure~\ref{fig:beans}(a) being mistaken for the background. These points can be readily filled by applying a dilation or closing operation onto the resulting segmentation(e.g.~\cite{najman2013mathematical,serra2012mathematical,costa2000shape}).

\begin{figure}
  \centering
     \subfloat[]{\includegraphics[width=0.45\textwidth]{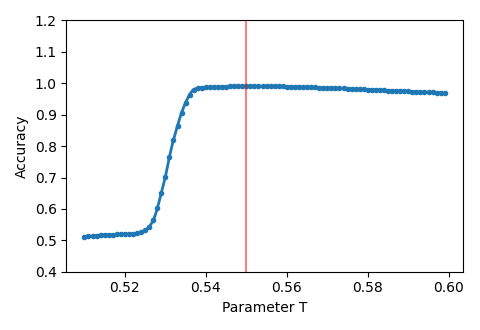}}
     \hspace{0.2cm}
     \subfloat[]{\includegraphics[width=0.45\textwidth]{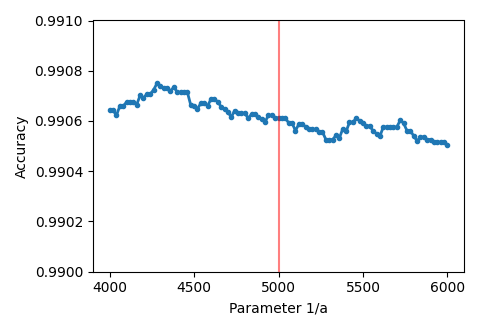}}
     
     \vspace{0.5cm}
     
     \subfloat[]{\includegraphics[width=0.45\textwidth]{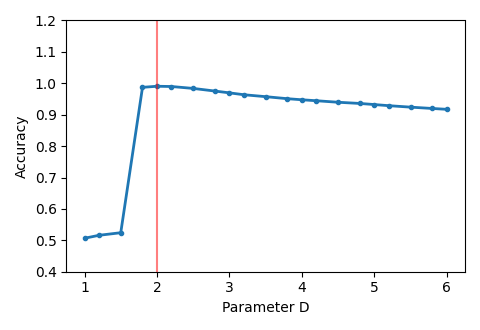}}
     \hspace{0.2cm}
     \subfloat[]{\includegraphics[width=0.45\textwidth]{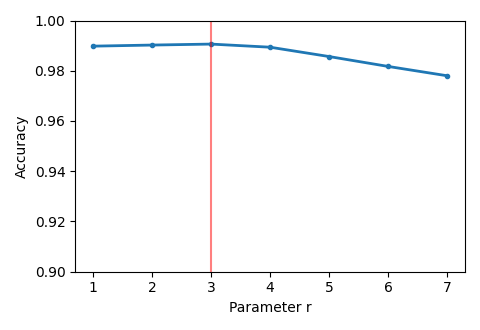}}
     
   \caption{Experiments estimating the accuracy obtained for the beans segmentation, considering prototype points within the objects, respective to independent variations around the identified optimal configuration (indicated by the red line).  Observe the different ranges along the $y-$axis.  The accuracy can be observed to be markedly stable for all parameters. In the specific case of the $D$ parameter, shown in (c), an abrupt can be observed at the left-hand side of the optimal point.}
  \label{fig:beans_Acc_obj}
\end{figure}

Given that the results above have been obtained for an optimized version of the parameters, it becomes particularly important to investigate how critical the performance depends on the choice of parameters. In order to better understand this issue, segmentations have been performed for parameter configurations around that which has been identified as optimal. Figure~\ref{fig:beans_Acc_obj} presents the accuracy of the segmentation for parameter values distributed around the respective optimal configuration, respectively to each of the four involved parameters.

Except for the parameter $D$, shown in Figure~\ref{fig:beans_Acc_obj}(c), the accuracy can be observed to be substantially robust to variations around the respective optimal configuration. In the particular case of the parameter $D$, an abrupt variation of accuracy resulted at the left-hand side of the optimal point. However, this transition can be easily verified and avoided while choosing the parameter configuration because of its strong impact on the respectively observed accuracy variation.

\begin{figure}
  \centering
     \subfloat[]{\includegraphics[width=0.45\textwidth]{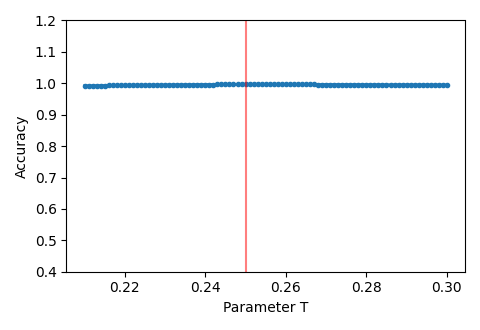}}
     \hspace{0.2cm}
     \subfloat[]{\includegraphics[width=0.45\textwidth]{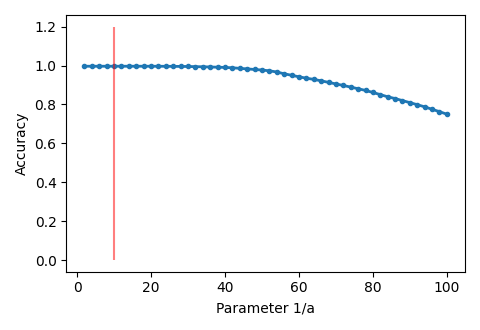}}
     
     \vspace{0.5cm}
     
     \subfloat[]{\includegraphics[width=0.45\textwidth]{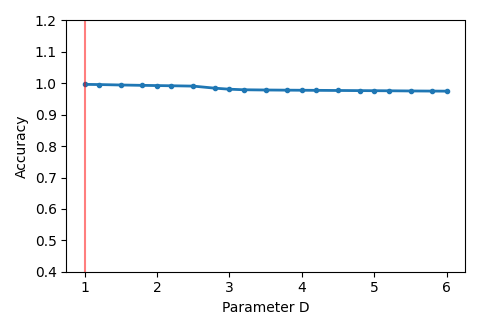}}
     \hspace{0.2cm}
     \subfloat[]{\includegraphics[width=0.45\textwidth]{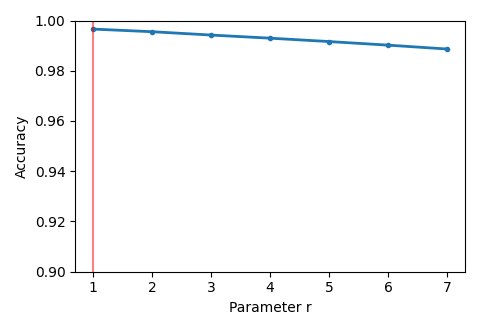}}
     
   \caption{Results of the experiments estimating the accuracy obtained for the beans segmentation, considering a single prototype point sampled from the background of the image, respective to variations around the identified optimal configuration (indicated by the red line). The accuracy can be observed to be markedly stable for all parameters. }
  \label{fig:beans_Acc_back}
\end{figure}

Figure~\ref{fig:beans_Acc_back} presents the analysis of the effect of parameter configurations respective to the beans image segmentation considering a single prototype point sampled from the respective background region.  

As a consequence of the background being substantially more homogeneous than the beans, an even smaller impact can be observed for variations around the optimal configuration of each of the four parameters.

\begin{figure}
  \centering
     \subfloat[]{\includegraphics[width=0.4\textwidth]{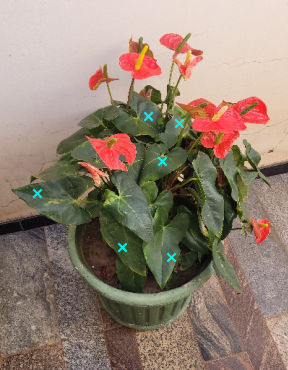}}
     \hspace{0.7cm}
     \subfloat[]{\fbox{\includegraphics[width=0.385\textwidth]{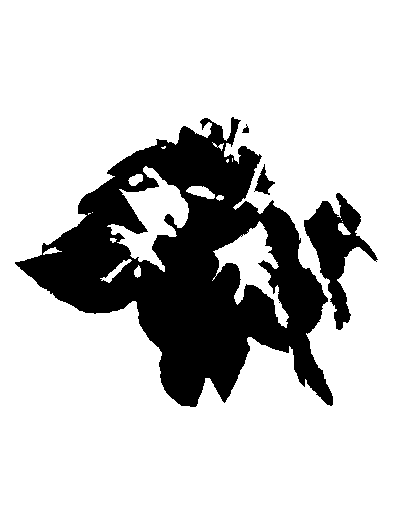}}}
     
     \vspace{0.9cm}
     
     \subfloat[]{\fbox{\includegraphics[width=0.385\textwidth]{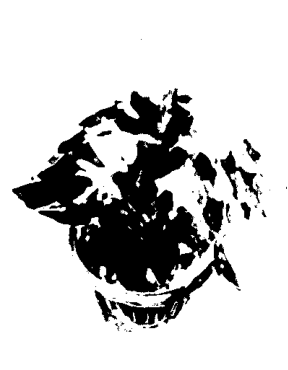}}}
     \hspace{0.7cm}
     \subfloat[]{\includegraphics[width=0.4\textwidth]{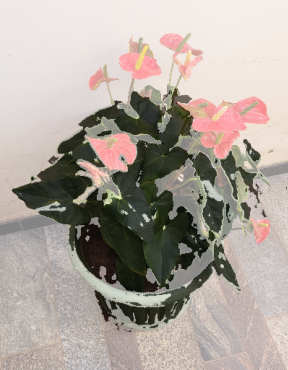}}
     
   \caption{Illustration of the potential of the CS-MNN methodology to segment a real-world image of a vase, with the region of interest (leaves) sampled in terms of six respective prototypes (shown in green).  The adopted parameter configuration was $D = 3$,  $r = 3$,  $T = 0.55$, and $a = 1/300.0$, which allowed an accuracy of $BA = 90.95$ \%.}
  \label{fig:vaso}
\end{figure}

Since the beans image in Figure~\ref{fig:beans} appear against a substantially uniform background, it becomes interesting to consider other images with more objects and a more cluttered background, such as the HSV vase image shown in Figure~\ref{fig:vaso}(a). In the following example, we will aim at segmenting the green leaves from the other portions of the vase, as well as from an intricate floor and a wall with varying luminosity. Observe the great variation of the points within the region of interest (leaves), presenting diverse hues, intensities, reflections, and shades, as well as additional structures (e.g. yellow veins). As a consequence, the segmentation of the leaves in this image by using computational approaches can be understood as representing a substantial challenge. 

Six prototype points were manually chosen for representing the leaves, which are depicted as green $\times$'s in Figure~\ref{fig:vaso}(a).

The gold standard mask, obtained manually, is shown in Figure~\ref{fig:vaso}(b). Observe that this mask corresponds to the green leaves of the anthurium plant. The segmentation obtained for the optimal parameter configuration is shown in Figure~\ref{fig:vaso}(c), while its superimposition onto the original image is presented in (d). Other than for the portion of the earth and the front of the vase being understood as leaves, the segmentation accuracy is substantially high.

\begin{figure}
  \centering
     \subfloat[]{\includegraphics[width=0.45\textwidth]{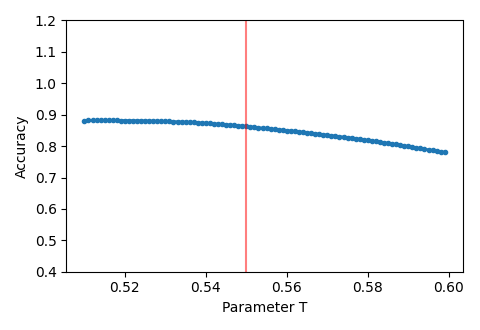}}
     \hspace{0.2cm}
     \subfloat[]{\includegraphics[width=0.45\textwidth]{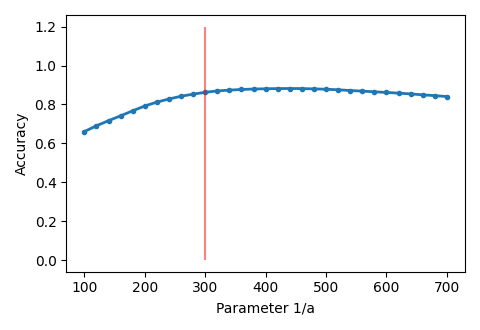}}
     
     \vspace{0.5cm}
     
     \subfloat[]{\includegraphics[width=0.45\textwidth]{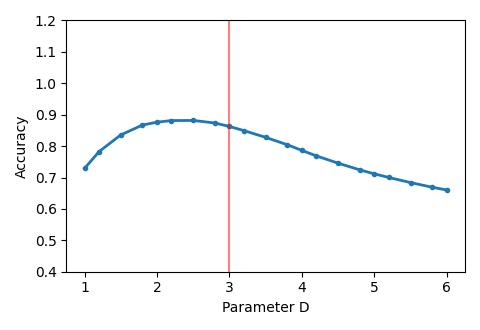}}
     \hspace{0.2cm}
     \subfloat[]{\includegraphics[width=0.45\textwidth]{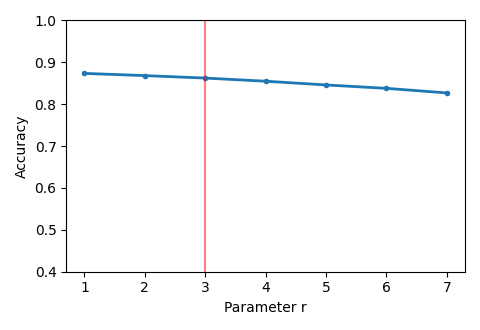}}
     
   \caption{Plots summarizing the experiments quantifying the accuracy obtained for the vase image segmentation, considering prototype points within the objects, respective to independent variations around the identified optimal configuration (indicated by the red line). Observe the different ranges along the $y-$axis. The accuracy can be observed to be markedly stable respectively to variations in all parameters. }
  \label{fig:vaso_Acc}
\end{figure}

Figure~\ref{fig:vaso_Acc} illustrates the influence of each of the four parameters (i.e.~$T$, $1/a$, $D$, and $r$) on the accuracy of the segmentation of the leaves in the vase image. Each of the plots shows the accuracy in terms of parameter values around the optimal respective configuration (shown by the red bar). Interestingly, as in the case of the beans image, the effect of the involved parameters again resulted relatively minor and even smoother, with small variations of accuracy being observed as a consequence of respective displacements. Observe that, though the maximum accuracy would have been obtained for $D=2$, instead of for $D=3$, this is explained by the fact that the implemented accuracy optimization took into account all the 4 parameter dimensions simultaneously.

\begin{figure}
  \centering
     \subfloat[]{\includegraphics[width=0.3\textwidth]{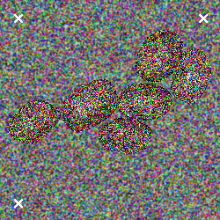}}
     \hspace{0.2cm}
     \subfloat[]{\includegraphics[width=0.3\textwidth]{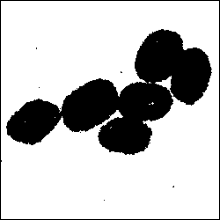}}
     \hspace{0.2cm}
     \subfloat[]{\includegraphics[width=0.3\textwidth]{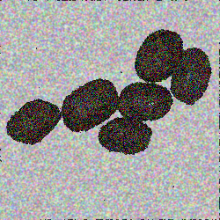}}
     
   \caption{Example of segmentation of a highly modified version of the HSV beans image in Fig.~\ref{fig:beans}(a) by using $D = 1$, $r = 2$, $T = 0.20$, and $a = 1/10$. Except for a few misclassified points and some jaggedness along the beans border, the resulting segmentation can be found to be highly accurate. The three chosen prototype points are shown as white $\times$'s.}
  \label{fig:beans_noise}
\end{figure}

As another example of the potential of the CS-MNN approach, it is now applied to the image shown in Figure~\ref{fig:beans_noise}(a), which corresponds to the beans image in Figure~\ref{fig:beans_noise} extensively modified by adding a random color noise field, to a point where the beans can hardly be identified by human inspection. The respective segmentation, shown in Figure~\ref{fig:beans_noise}(b), substantiates the marked stability of the CS-MNN approach: all beans have been properly segmented, except for some jaggedness in the beans borders and a few isolated points incorrectly classified within the objects and background.

\begin{figure}
  \centering
     \subfloat[]{\includegraphics[width=0.3\textwidth]{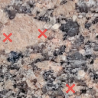}}
     \hspace{0.2cm}
     \subfloat[]{\includegraphics[width=0.3\textwidth]{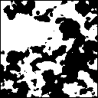}}
     \hspace{0.2cm}
     \subfloat[]{\includegraphics[width=0.3\textwidth]{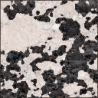}}
     
   \caption{Example of segmentation of an HSV granite image by using $D = 3$, $r = 2$, $T = 0.65$, and $a = 1/2,000$.
   The high accuracy of the obtained segmentation can be better appreciated by looking at the above figure from further away than usual, as this allows less intense adaptation to contrast.}
  \label{fig:marmore}
\end{figure}

In order to illustrate the performance of the CS-MNN methodology respectively to an even more complex image, it has been applied to the HSV image of granite stone shown in Figure~\ref{fig:marmore}(a). The objective here is to segment the regions that do not correspond to the cream-colored portions of the granite, which have been manually sampled in terms of the three prototype points shown as red $\times$'s in Figure~\ref{fig:marmore}(a).

The parameters were interactively set in order to yield as much as accurate results (by visual comparison with the original images). Given the extreme complexity of this type of image, no gold standard reference could have been obtained. However, a visual comparison between the original image in Figure~\ref{fig:marmore}(a) and the segmentation result shown in (b) readily indicates a marked adherence of the segmented parts (shown in black) as corresponding to the sought regions. The superimposition of the segmented and original images is shown in Figure~\ref{fig:marmore}(c). 

\begin{figure}
  \centering
     \subfloat[]{\includegraphics[width=0.3\textwidth]{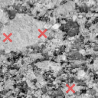}}
     \hspace{0.2cm}
     \subfloat[]{\includegraphics[width=0.3\textwidth]{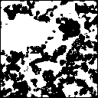}}
     \hspace{0.2cm}
     \subfloat[]{\includegraphics[width=0.3\textwidth]{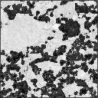}}
     
   \caption{Segmentation of the image in Fig.~\ref{fig:marmore}(a) while considering its respective gray-level version, with parameter configuration $D=3$, $r=2$, $T=0.7$, and $a=1/2,000$.  As with the previous figure, the quality of the obtained segmentation can be better appreciated by looking from a distance larger than usual.}
  \label{fig:marmore_gray}
\end{figure}

In order to illustrate the effect of taking a gray-level version of a color image, in the following we address the segmentation of the same previous granite example, but now taking into account its converted gray-level version. The same parameter configuration used in the previous example has been applied also in this case. The gray-scale of the original image is shown in Figure~\ref{fig:marmore_gray}(a), and its respective segmentation is depicted in (b). The superposition of these two images is shown in (c).

Interestingly, the segmentation regions resulted in a substantially more detailed identification of the non-cream-colored parts of the granite. This result has been a direct consequence of the lack of information about the color properties, in which case the segmentation became mostly driven by the varying gray-levels. In a sense, the results obtained for the HSV and gray-level versions of the granite image can be understood, at least for this type of image, to provide valid results characterized by distinct spatial scales, with a higher level of detail being provided by the gray-level segmentation. These two images can then be combined in order to obtain a more complete representation and analysis.

\subsection{Comparison with a Deep Learning Approach I: Granite Images}

In this section, we report on experiments that have been performed in order to compare the segmentation of a specific type of image (granite stones) as obtained by using the CS-MNN as well as a deep learning implementation based on the U-Net architecture.

Granite stone images have been chosen because of their particularly intricate coarse-grained structure, with medium to low contrast clusters with varying colors and gray-level intensities, as well as presence of simultaneous spatial scales. Granite stone (e.g.~\cite{le2005igneous}) is a composition of feldspar, quartz, and biotite mica. Porphyritic granite stones, such as those in the adopted images, are characterized by a marked variance of domains. These multiple geometric scales allow a more systematic evaluation of methods for respective segmentation. In addition, in the case of the granite in the adopted images, the separation among the domains is often graded, being characterized by progressive variation of color and/or intensity which also make the respective segmentation particularly challenging. As a consequence of the above observed features, accurate human-assisted segmentation of these images becomes virtually impossible. At the same time, and for the same reasons, the adopted images constitute a particularly difficult segmentation problem.

As a consequence of the specific choice of images, deep learning architecture, and respective parameter configurations, the results described in the following are necessarily limited to these particular choices and ae not direct or immediately extensible to other possible situations. Indeed, the objective of performing this comparison is not to reach any decisive conclusions regarding these two approaches, but rather to illustrate some of their main interesting features when applied to the particular type of chosen images.

A data set incorporating 45 HSV granite images, each with dimension $128\times128$, has been obtained. The segmentation objective is henceforth understood to identify the regions of the granite images that do not correspond to the cream-colored parts.

As a consequence of the intrinsic complexity of these images, it becomes practically impossible to obtain, through human manipulation, accurate respective gold standard references. At the same time, the markedly accurate results obtained for this type of image reported in Subsection~\ref{sec:Preliminary} suggest that the segmentation achieved by using the coincidence-based similarity could be considered as gold standards for the granite images. In this work, the gold standard images therefore correspond to the CS-MNN results with $D=3$, $r=2$, $T=0.65$, and $a=1/2,000$. This choice, therefore, means that the comparison experiment will mainly concern the quantification of the capability of the chosen deep learning implementation in reproducing the results preliminary obtained by the coincidence-based similarity.

The specific deep learning implementation described in Subsection~\ref{sec:deepsegmentation} is henceforth considered for the following comparison. In particular, 40 images were selected for the training stage, leaving 5 images to be tested afterward. The U-Net architecture was trained using batch size 4 along 20 epochs.

\begin{figure}
  \centering

\emph{Image 1}:   \hspace{0.2cm}  
  \includegraphics[width=0.7\textwidth]{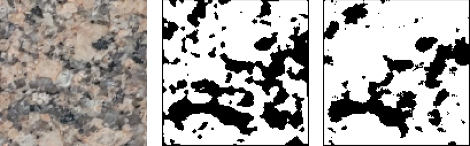}

\vspace{0.3cm}

\emph{Image 2}:   \hspace{0.2cm}
     \includegraphics[width=0.7\textwidth]{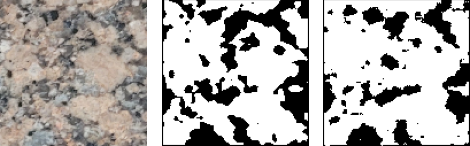}

\vspace{0.3cm}

\emph{Image 3}:   \hspace{0.2cm}
     \includegraphics[width=0.7\textwidth]{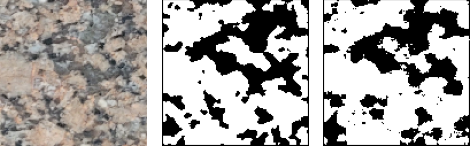}

\vspace{0.3cm}

\emph{Image 4}:   \hspace{0.2cm}
     \includegraphics[width=0.7\textwidth]{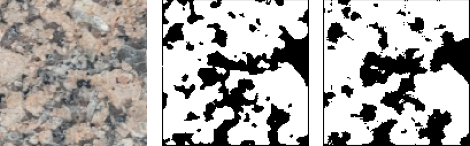}

\vspace{0.3cm}

\emph{Image 5}:   \hspace{0.2cm}
     \includegraphics[width=0.7\textwidth]{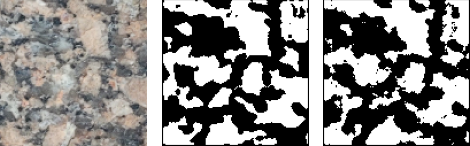}

   \caption{The five test granite HSV images and their respective segmentation by using the CS-MNN and deep learning approaches while considering specific parameter configurations. Though the results are moderately similar, the images obtained by using the coincidence-based approach tended to be smoother, and involved fewer isolated points, allowing for better accuracy when visually comparatively with the original granite images shown in the first column of this figure.}
  \label{fig:deep}
\end{figure}

Figure~\ref{fig:deep} presents the results obtained for the 5 test granite images by using the coincidence-based and deep learning approaches.

These results indicate that moderately similar segmentations have been obtained by both methods. However, the deep learning results are characterized by a larger number of isolated points, as well as less smooth regions and reduced accuracy in the identification of regions of interest. The accuracy values obtained for the deep learning approach are given in Table~\ref{tab:deepaccuracy}.

\begin{table}[]
\centering
\caption{Accuracy values obtained for the deep learning-based segmentation of the five granite test images.}\label{tab:deepaccuracy}
\begin{tabular}{|l|c|}
\hline
Image id  &  Accuracy (BA) \\  \hline
\emph{Image 1} & 82.53 \% \\
\emph{Image 2} & 86.56 \% \\
\emph{Image 3} & 84.85 \% \\
\emph{Image 4} & 83.37 \% \\
\emph{Image 5} & 88.01 \% \\ \hline
\end{tabular}
\end{table}

The best deep learning result was obtained for image 5, while images 1 and 3 implied more substantial challenges to that methodology. When compared to the accuracy values obtained in the cases presented in Subsection~\ref{sec:Preliminary}, the values in Table~\ref{tab:deepaccuracy} can be understood to be relatively moderate.

Though a relatively small number of sample images have been chosen for training the deep learning architecture, they actually provide hundreds of thousands of prototype points, which are the basic training units of the coincidence-similarity approach. Therefore, when considered from this perspective, a substantial number of training samples at comparable granularity have actually been considered while training the deep learning approach.

The experiments described above indicate that the deep learning approach allowed moderate accuracy, depending on the set of images, parameter configurations, and comparison method. At the same time, the time and hardware resources implied by the deep learning approach are substantially greater than those demanded by the coincidence-based methodology, which requires only 3 elementary multiset neurons implying a total of $(3)(13)=39$ parameters (weights) (for $r=2$) instead of the 6,502,786 total parameters (4,658,882 trainable and 1,843,904 non-trainable) required by the adopted deep learning configuration.

While the coincidence-based approach used 3 prototype points for each image, the deep learning can proceed in a fully automated manner after being trained. Another interesting comparison aspect concerns the fact that the way in which the multiset method operates can be fully characterized and understood in qualitative terms. At the same time, the identification of how the deep learning approach solved this specific segmentation problem consists of a relatively challenging issue.

Table~\ref{tab:comparison} summarizes some of the main features allowed by each of the two methodologies considered in this work.

\begin{table}
\caption{Table with the main characteristics of image segmentation by deep learning and coincidence similarity-based multiset neuronal networks.}\label{tab:comparison}
\centering
\begin{tabular}{l|c|c}

                                & \textbf{Deep Learning} & \textbf{ CS-MNN}  \\ 
\hline\hline
Complexity of Neurons           & Average       & Average   \\ \hline
Flexibility of Neurons          & Moderate      & High      \\ \hline
Training samples per image      & 16384         & 3         \\ \hline
Number of neurons               & tens of thousands     & 3 \\
\hline
Number of layers                & very large         & 1         \\ \hline
Explainability                  & challenging   &  straightforward \\ \hline
Generalization                  & good          & good      \\ \hline
Accuracy (test stage)           & $\left[ 82.64 \pm 3.52\right]$\%     & 
\begin{tabular}[c]{@{}l@{}}
$100\%$ (gold standard) 
\end{tabular}\\ 
\hline\hline
\end{tabular}
\end{table}

As discussed in~\cite{costa2023similarity,costa2023multiset}, Jaccard similarity neurons are more flexible than inner-product-based neurons commonly adopted in convolutional networks, as combinations of a handful of multiset neurons are capable of implementing highly intricate polyhedric decision regions.

Solutions obtained by the CS-MNN approach are intrinsically explainable in terms of the logical union of three basic similarity comparisons with specific sets of weights defining respective diamond regions in the feature space. At the same time, it constitutes a substantially more demanding task to extract knowledge from the typically intricate architectures, involving a large number of neurons, underlying deep learning approaches.

Both approaches have been understood to have good comparable generalization capabilities, in the sense that the same weights and configuration parameters allowed good results to be obtained over the whole set of granite images. Further evaluation of the generalization capabilities of these approaches could also consider the application of the trained configurations to other types of stone, or even objects.

The accuracy obtained for the CS-MNN is $100\%$, as a consequence of taking its results as the gold standard. As already observed, the performed comparison was thus primarily aimed at verifying to which extent the deep learning methodology allowed the replication of the accurate results obtained by the CS-MNN. A complementary approach is reported in the following sections, respectively to synthetic images.

\subsection{Comparison with a Deep Learning Approach II: Synthetic Images}

In the previous section, the CS-MNN methodology was compared with the adopted deep learning approach respectively to complex granite images while taking the results of the former as the gold standard. Therefore, that comparison ultimately investigated how well the deep learning approach could replicate the CS-MNN results for that considered set of 45 real-world images. 

Reported in the present section is a complementary and more typical comparison experiment, between the same two previous methods, but now using complex images synthetically generated by a respective methodology described as follows, which allows generating of a virtually unlimited number of images together with the respective gold standards, from which the images to be segmented are actually obtained.

The considered synthetic image generation included the following steps: (a) a random field of $P$ discrete points is generated in a $N \times N$ image in a uniformly manner, but satisfying an exclusion radius $e$; (b) this field is convolved with a circularly symmetric Gaussian kernel with the respective dispersion quantified by the parameter $\sigma$; (c) the resulting image is thresholded by a value $R$, yielding a respective binary image, which is henceforth considered as the respective gold standard; and (d) each of the two regions of the mask is filled with textures of different spatial scales, corresponding to random fields of $N_{obj}$ and $N_{bj}$; and (d) each of the two regions of the mask is filled with textures of different spatial scales, corresponding to random fields of $N_{obj}$ and $N_{bck}$ points, respectively, smoothed by Gaussian kernels with $\sigma_{obj}$ and $\sigma{bck}$ for the object interior and background, respectively. Additionally, the object and background are generated with different mean values $\mu_{obj}$ and $\mu_{bck}$.

In addition to the images to be segmented, the CS-MNN method requires a set of prototype points. We obtained these points by choosing a subset of $Q$ points from the $P$ points that were used to create the gold standard. These points are at least $r$ pixels away from the border (where $r$ is the window size parameter's radius, see Section~\ref{sec:CS-MNN}). This has been addressed by eroding the gold standard $r$ times and selecting the $Q<P$ points that remained within it. This margin for the prototype points prevents background pixels from being included in the weights that define the multiset neurons for each prototype. We adopted $Q=5$ points for all subsequent experiments.

The above methodology for generating and preparing the images to be segmented by the two considered methodologies, therefore, provides a fully automated manner of obtaining a virtually unlimited number of training and test images, together with their respective gold standard and randomly chosen prototype points. In this manner, it becomes possible to train the deep learning methodology while considering a more substantial number of training images. The practical limitation to the number of considered images is then established by the computer time and resources required for training the deep learning architecture.

To evaluate the performance of the proposed CS-MNN method, we performed an experiment using synthetic images generated with the following parameters: image size $N = 128$, exclusion radius $e=6$, number of initial points $P = 60$, the standard deviation of the Gaussian used to smooth these points $\sigma = 5$ pixels, threshold applied to the smoothed regions $R = .3$, size of the set of prototype points $Q = 5$, mean gray-level of the objects $\mu_{obj} = 165$, mean gray-level of background $\mu_{bck} = 65$, standard deviation used to smooth the object texture $\sigma_{obj}$ = 0.7 pixel, and standard deviation applied to the background  $\sigma_{bck}$ = 1 pixel. We set the CS-MNN parameters as follows: $D = 2$, $r = 4$, $T = 0.85$, and $a = 1/5000.0$. As indicated in~\cite{costa2023multiset, da2022supervised}, the gray levels in each widowed region around the prototype points were sorted in increasing order. We trained a U-Net model with a batch size of 10 for 30 epochs, using 170 synthetic images for training and 30 for testing.

\begin{figure}
  \centering
     \includegraphics[width=0.6\textwidth]{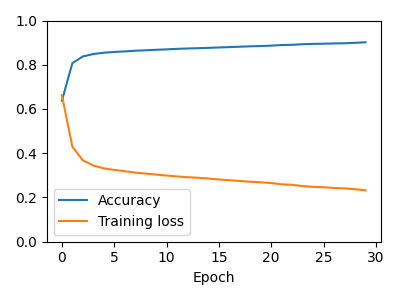}
   \caption{The accuracy and loss in terms of the training epochs obtained for the segmentation experiment are described in this section.}\label{fig:TrainingLoss}
\end{figure}

The values of accuracy and loss obtained along the training epochs in the case of the deep learning approach are depicted in Figure~\ref{fig:TrainingLoss}. The accuracy can be observed to undergo an abrupt increase, followed by a regime of slower increase.

\begin{figure}
  \centering

\emph{Image 1}:   \hspace{0.2cm}  
  \includegraphics[width=0.75\textwidth]{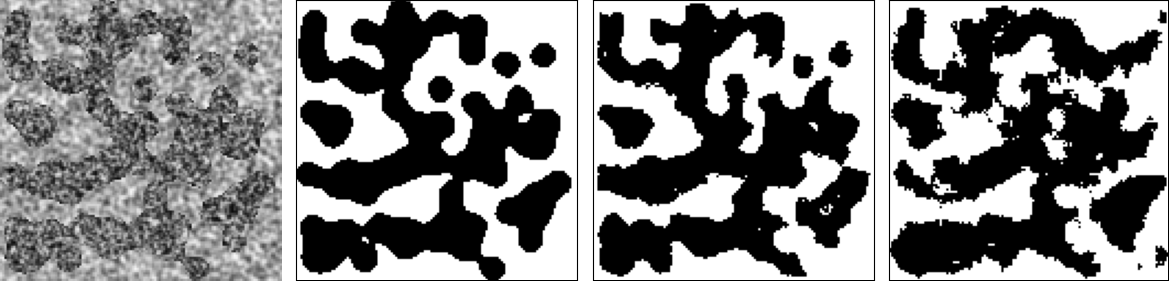}

\vspace{0.3cm}

\emph{Image 2}:   \hspace{0.2cm}
     \includegraphics[width=0.75\textwidth]{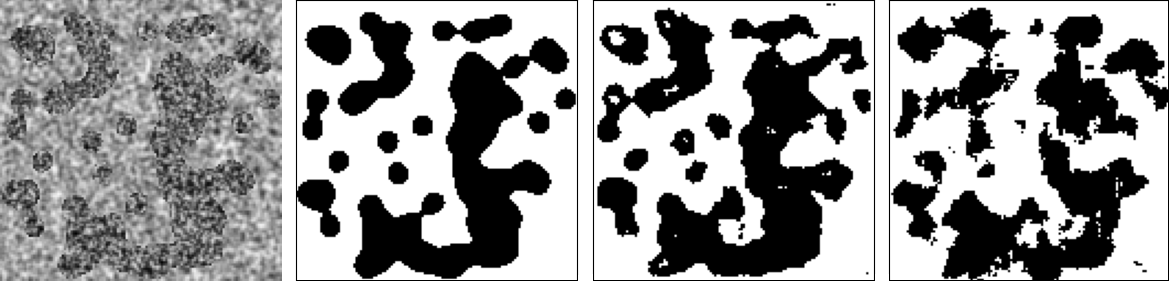}

\vspace{0.3cm}

\emph{Image 3}:   \hspace{0.2cm}
     \includegraphics[width=0.75\textwidth]{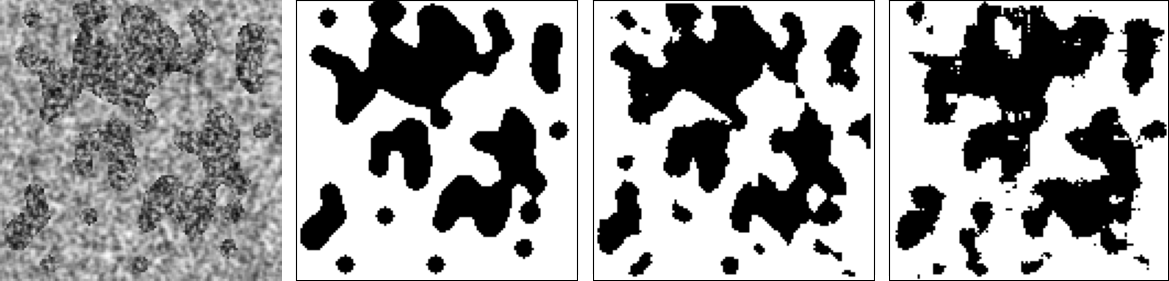}

\vspace{0.3cm}

\emph{Image 4}:   \hspace{0.2cm}
     \includegraphics[width=0.75\textwidth]{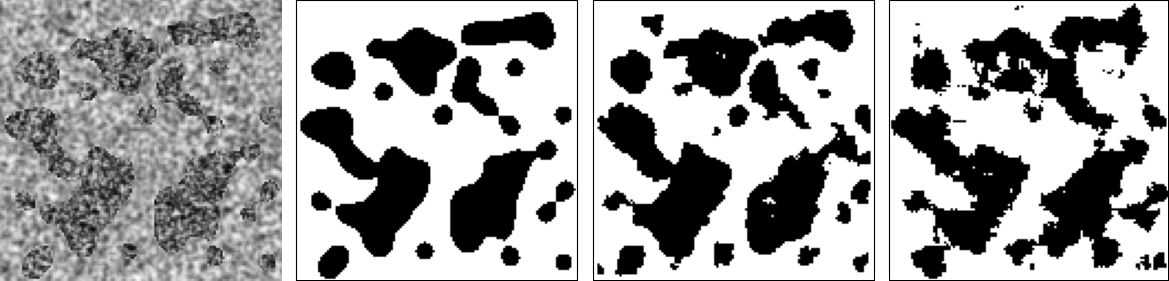}

\vspace{0.3cm}

\emph{Image 5}:   \hspace{0.2cm}
     \includegraphics[width=0.75\textwidth]{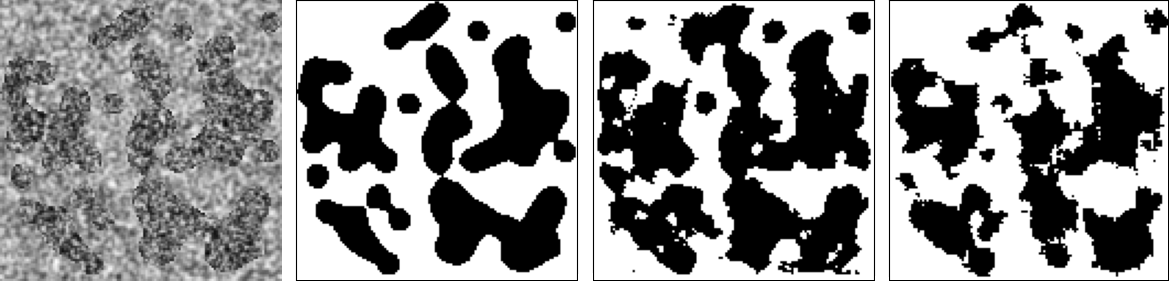}
   \caption{Examples of synthetic images (first column), respective gold standards (second column), and the segmentation results obtained by using the CS-MNN (third column) and deep learning  (fourth column) approaches.}
  \label{fig:synth}
\end{figure}

Figure~\ref{fig:synth} shows some of the synthetic images (first column), respective gold standards (second column), as well as the segmentation results obtained by using the CS-MNN (third column) and adopting a specific deep learning approach (fourth column). Recall that the gold standards were obtained automatically, independently of any of the two approaches. The obtained balanced accuracy values are presented in Table~\ref{tab:seg_accuracy}.

\begin{table}[]
\centering
\caption{Accuracy values obtained for the deep learning-based segmentation of the five synthetic test images in Fig.~\ref{fig:synth}.}\label{tab:seg_accuracy}
\begin{tabular}{|l|c|c|}
\hline
Image id  & Deep Learning Accuracy (BA) &  CS-MNN Accuracy (BA)\\ \hline
\emph{Image 1} & 84.42 \% & 92.15 \% \\
\emph{Image 2} & 84.48 \% & 92.16 \% \\
\emph{Image 3} & 87.41 \% & 92.47 \% \\
\emph{Image 4} & 88.53 \% & 91.91 \% \\
\emph{Image 5} & 86.21 \% & 88.69 \% \\ \hline
\end{tabular}
\end{table}

\begin{figure}
  \centering
     \includegraphics[width=0.6\textwidth]{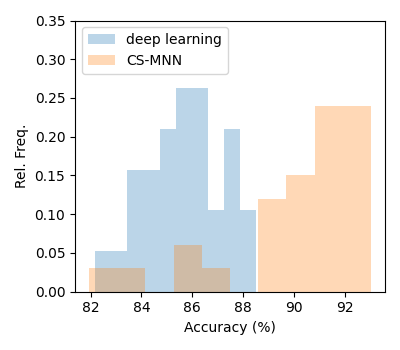}
   \caption{Histograms of the relative frequency of the accuracy values obtained for the two considered supervised segmentation methodologies respectively to the database of synthetic images adopted in this section.}\label{fig:accuray_distribution}
\end{figure}

Figure~\ref{fig:accuray_distribution} presents the histograms of the relative frequency of the accuracy values obtained for the two considered supervised segmentation approaches. The CS-MNN allowed substantially enhanced performance, except for a few cases presenting accuracy comparable to those obtained for the deep learning approach. It should be observed that, in the case of the CS-MNN, the prototype points have been obtained in a uniformly random manner among the $P$ initial points. Therefore, further improved accuracy values can be expected when the prototype points are chosen in a more objective manner.

Table~\ref{tab:comparison2} summarizes the main features allowed by each of the two supervised segmentation methodologies adopted in this work, which are similar to those shown in Table~\ref{tab:comparison}, except for the number of neurons in the CS-MNN and overall accuracy values obtained for each of the two methodologies.

\begin{table}
\caption{Table with the main characteristics of image segmentation by deep learning and coincidence similarity-based multiset neuronal networks.}\label{tab:comparison2}
\centering
\begin{tabular}{l|c|c}

                                & \textbf{Deep Learning} & \textbf{ CS-MNN}  \\ 
\hline\hline
Complexity of Neurons           & Average       & Average   \\ \hline
Flexibility of Neurons          & Moderate      & High      \\ \hline
Training samples per image      & 16384         & 5         \\ \hline
Number of neurons               & tens of thousands     & 5 \\
\hline
Number of layers                & very large         & 1         \\ \hline
Explainability                  & challenging   &  straightforward \\ \hline
Generalization                  & good          & good      \\ \hline
Accuracy (170 training images)  & $\left[ 85.67 \pm 1.48\right]$\%     &  $\left[ 90.10 \pm 2.68\right]$\% \\ \hline
Accuracy (4,000 training images)  & $\left[ 87.67 \pm 1.34\right]$\%     &  $\left[ 89.39 \pm 4.13\right]$\% \\

\hline\hline
\end{tabular}
\end{table}

In order to complement the analysis with the synthetic images, we repeated the above experiments with the same configuration for image generation, but using 4,500 images, 4,000 for training, and 500 for testing. Now, The U-Net architecture was trained using batch size 50 along 50 epochs.

\begin{figure}
  \centering
     \includegraphics[width=0.6\textwidth]{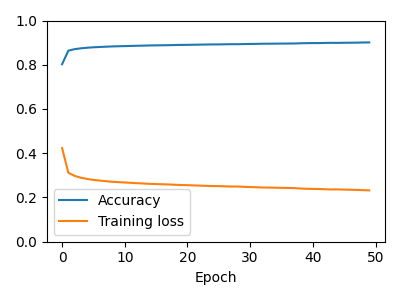}
   \caption{The accuracy and loss in terms of the training epochs obtained for the segmentation experiment are described in this section, but considering 4,500 synthetic images.}\label{fig:TrainingLoss2}
\end{figure}

Figure~\ref{fig:TrainingLoss2} presents the accuracy and loss obtained for the larger set of images. The overall shape of the accuracy curve is similar as before, including an abrupt increase at its beginning, followed by slower growth.

\begin{figure}
  \centering

\emph{Image 1}:   \hspace{0.2cm}
  \includegraphics[width=0.75\textwidth]{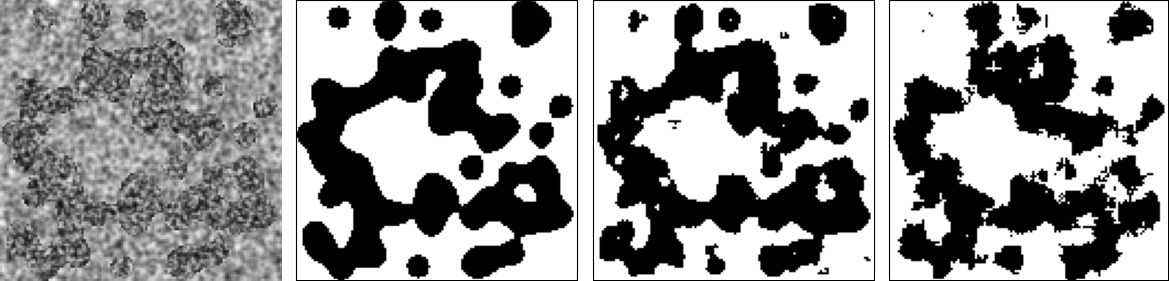}

\vspace{0.3cm}

\emph{Image 2}:   \hspace{0.2cm}
     \includegraphics[width=0.75\textwidth]{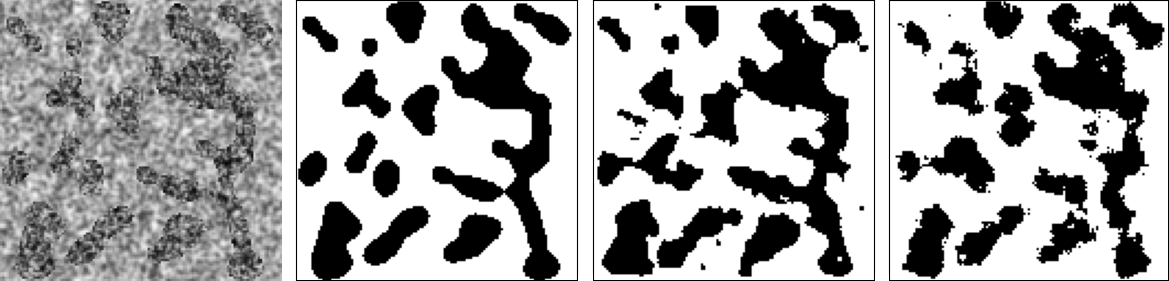}

\vspace{0.3cm}

\emph{Image 3}:   \hspace{0.2cm}
     \includegraphics[width=0.75\textwidth]{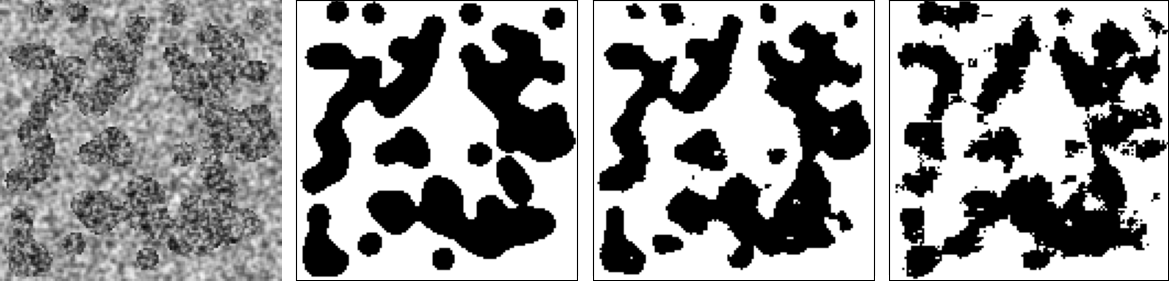}

\vspace{0.3cm}

\emph{Image 4}:   \hspace{0.2cm}
     \includegraphics[width=0.75\textwidth]{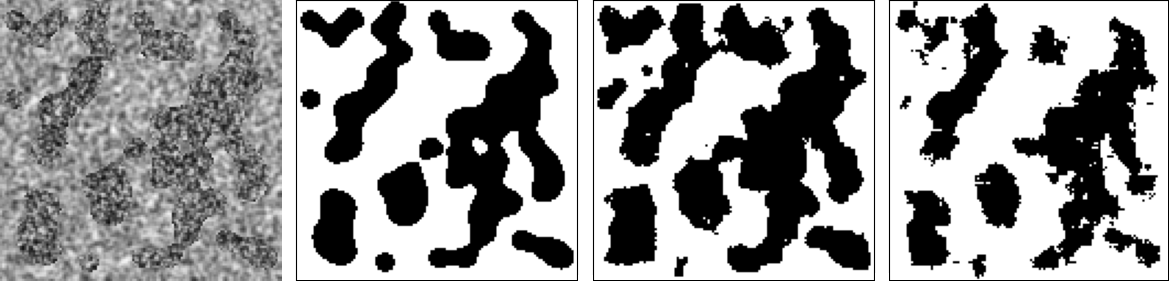}

\vspace{0.3cm}

\emph{Image 5}:   \hspace{0.2cm}
     \includegraphics[width=0.75\textwidth]{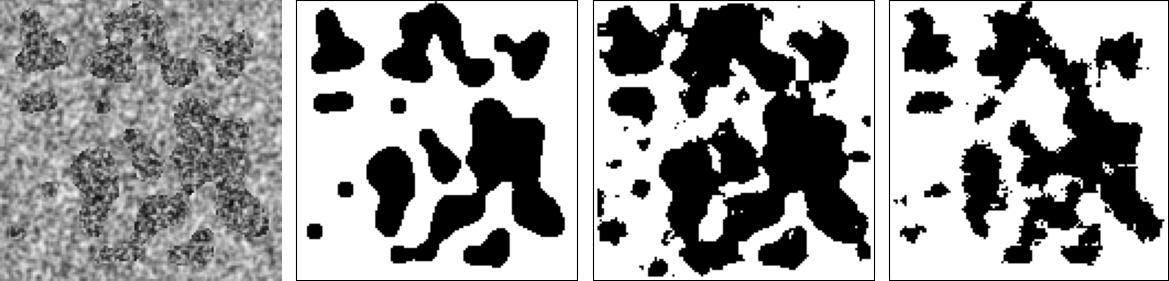}
   \caption{Examples of synthetic images (first column), respective gold standards (second column), and the segmentation results obtained by using the CS-MNN (third column) and deep learning (fourth column) approaches. The deep learning approach was trained with 4,000 synthetic images instead of the 170 synthetic images used for the training whose results are shown in Fig.~\ref{fig:synth}}
  \label{fig:synth2}
\end{figure}

Figure~\ref{fig:synth2} illustrates five synthetic images (first column), as well as their respective gold standards (second column), and segmented images obtained by using the CS-MNN (third column) and deep learning (fourth column), approaches, the latter considering 4,000 training images instead of the 170 adopted in the previous experiment. Table~\ref{tab:seg_accuracy2} shows the balanced accuracy values obtained for each of these images.

\begin{table}[]
\centering
\caption{Accuracy values obtained for the deep learning-based segmentation of the five synthetic test images in Fig.~\ref{fig:synth2}.}\label{tab:seg_accuracy2}
\begin{tabular}{|l|c|c|}
\hline
Image id & Deep Learning Accuracy (BA) & CS-MNN Accuracy (BA)\\ \hline
\emph{Image 1} & 86.49 \% & 92.26 \% \\
\emph{Image 2} & 88.10 \% & 90.33 \% \\
\emph{Image 3} & 84.09 \% & 91.87 \% \\
\emph{Image 4} & 86.92 \% & 90.66 \% \\
\emph{Image 5} & 88.43 \% & 87.91 \% \\ \hline
\end{tabular}
\end{table}

\begin{figure}
  \centering
     \includegraphics[width=0.6\textwidth]{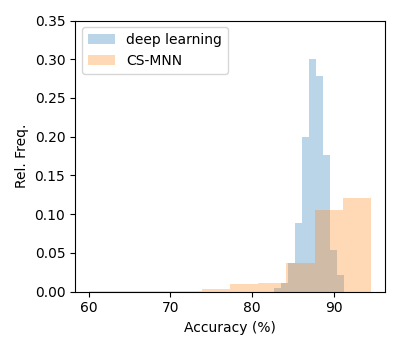}
   \caption{Histograms of the relative frequency of the accuracy values obtained for the two considered supervised segmentation methodologies respectively to experiment involving 4,500 synthetic images. The accuracy values obtained for the deep learning approach had their dispersion reduced while mostly preserving the same average.}\label{fig:accuray_distribution2}
\end{figure}

The histograms of relative frequencies obtained for the experiment with 4,5000 synthetic images is depicted in Figure~\ref{fig:accuray_distribution2}. It can be observed that, though the accuracy of the deep learning approach presented a reduction of dispersion, it still remains lower in the average than the accuracy observed for the CS-MNN methodology. The larger accuracy dispersion obtained for the latter methodology is a consequence of the automatic choice of prototype points, which tends to yield less accurate results. However, good CS-MNN performance has been obtained even in this situation.

\begin{figure}
  \centering
     \includegraphics[width=0.6\textwidth]{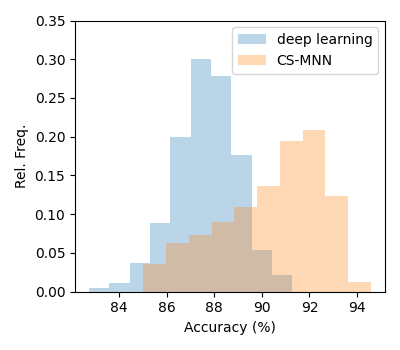}
   \caption{New version of the accuracy histograms in Fig.~\ref{fig:accuray_distribution2}, obtained by choosing new sets of five points in the cases in which the CS-MNN accuracy was smaller than $85\%$. The results obtained from the deep learning approach remained unchanged.}\label{fig:accuray_distribution3}
\end{figure}

In order to confirm that the few cases in which the CS-MNN led to relatively low balanced accuracy values were a consequence of the automated choice of prototype points, the whole set of synthetic images was segmented again in a different manner. More specifically, the images leading to balanced accuracy smaller than $85\%$ were identified, and a new respective set of five prototype points were randomly re-drawn. Though not yet being as effective of the human-assisted choices of prototype points, which is virtually impossible given the large number of considered images, this approach led to a substantial improvement of the segmentation accuracy in the case of the CS-MNN methodology, as illustrated in Figure~\ref{fig:accuray_distribution3}. The so-obtained results fully corroborate the previous interpretation of the cause of the few low accuracy cases resulting from the previous experiment.

\section{Concluding Remarks} \label{sec:conclusion}

The task of identifying regions of interest from 2D gray-level or color images, typically known as image segmentation, constitutes a substantial challenge that has motivated a considerable number of interesting approaches. More recently, developments in deep learning have made it possible to perform image segmentation effectively and in a more general way, thus becoming a reference method.

Also recently, an alternative method for supervised image segmentation (here called CS-MNN) based on the application of the coincidence multiset operation has been described, which has yielded encouraging results characterized by high accuracy while relying on minimal computational resources and training time.  These interesting properties are ultimately inherited from the coincidence similarity index which constitutes the basis for the considered multiset neuronal networks. 

After describing the two adopted image segmentation methods, the present work has developed two experiments aimed at comparing the CS-MNN and deep learning approaches, which mainly aim at illustrating their respective interesting features.

The first reported comparison experiment considered a specific data set containing 45 colors (HSV) intricate images of granite stone. Since it is practically impossible to manually obtain gold standards for these images, and considering the high accuracy allowed by the multiset approach respectively for this type of images, the segmentation results yielded by the latter approach have been adopted as gold standard references. This means that the comparison experiment becomes mainly a study of how well the adopted deep learning architecture can replicate the performance of the CS-MNN respectively to these particular gold standards.

The results observed in this first experiment indicate that deep learning achieves moderate accuracy in segmentation compared to the CS-MNN method, which requires significantly fewer computational resources.

Two additional experiments were performed respectively to a larger set of synthetic images (200 and 4.500 images) and associated gold standards obtained by a suggested generating methodology, which also provided respective prototype points to be used in the CS-MNN method, therefore resulting in a completely automated obtaining a virtually unlimited number of images and gold standards. These images were characterized by containing objects against a background, having filling textures with distinct averages and dispersions. It should be observed that the described methodology to generate synthetic images can be applied as a subsidy to other systematic comparisons involving a large number of images with respective gold standards.

The results obtained for the segmentation of synthetic images were similar to those observed in the first experiment with granite images, with the deep learning approach leading to relatively accurate segmentation, while the CS-MNN methodology allowed further enhanced accuracy while requiring little computational resources.

It should be observed that the reported comparison, which is respective to the adopted specific data types and parameter configurations as well as other choices, is in no way intended to provide a conclusion on which of the two approaches is superior to the other in a more general context. Indeed, our discussion is limited to providing an illustration of the most interesting features of each of the two considered methodologies while referring only to the adopted specific types of images, parameter configurations, and other respective choices.

While the deep learning approach confirmed its potential and generality by segmenting the considered images in an accurate and fully automated manner (after the training stage), the multiset methodology is performed under human supervision, requiring a few points to be chosen for each of the images to be segmented. At the same time, the latter approach required few computational resources, involving only a few multiset neurons implemented in terms of respective coincidence similarity operations associated to each chosen prototype. Both approaches were characterized by marked generalization capabilities, as indicated by the overall good results obtained while using the same parameter and training configurations.

Additional related studies motivated by the reported concept, methods, and results, include the adoption of the recently introduced idea of quantification of similarity between a data element and \emph{a whole set of other elements}~\cite{costa2023toward, costa2023similarity} for obtaining a respective multiset network.

\section*{Acknowledgments}
Luciano da F. Costa thanks CNPq(grant no. 307085/2018-0) and FAPESP (grant 15/22308-2).

\bibliography{ref}
\bibliographystyle{unsrt}

\end{document}